\documentclass[a4paper]{article}

\usepackage{INTERSPEECH2021}
\usepackage{breakurl}
\usepackage[bookmarks=false]{hyperref}
\hypersetup{
   breaklinks=true,   
   colorlinks=true,   
   pdfusetitle=true,  
}
\usepackage{multirow}

\title{Improving Contextual Recognition of Rare Words\\with an Alternate Spelling Prediction Model}
\name{Jennifer Drexler Fox, Natalie Delworth}
\address{Rev.com}
\email{\{jennifer.drexler, natalie.delworth\}@rev.com}

\begin{document}

\maketitle

\begin{abstract}
Contextual ASR, which takes a list of bias terms as input along with audio, has drawn recent interest as ASR use becomes more widespread. We are releasing contextual biasing lists to accompany the Earnings21 dataset, creating a public benchmark for this task. We present baseline results on this benchmark using a pretrained end-to-end ASR model from the WeNet toolkit. We show results for shallow fusion contextual biasing applied to two different decoding algorithms. Our baseline results confirm observations that end-to-end models struggle in particular with words that are rarely or never seen during training, and that existing shallow fusion techniques do not adequately address this problem. We propose an alternate spelling prediction model that improves recall of rare words by 34.7\% relative and of out-of-vocabulary words by 97.2\% relative, compared to contextual biasing without alternate spellings. This model is conceptually similar to ones used in prior work, but is simpler to implement as it does not rely on either a pronunciation dictionary or an existing text-to-speech system.
\end{abstract}

\section{Introduction}
Contextual ASR is the task of biasing an ASR system towards a user-defined list of important words and phrases that are submitted along with the audio to be transcribed. In the literature, contextual ASR is most often studied by industry research labs as part of their work on virtual assistant devices with speech interfaces: it is important that these systems recognize the names of a user's contacts when that user wants to make a call or the names of artists in a user's music library when that user wants to hear a particular song. Because this task requires user-specific information, the majority of papers published on contextual ASR use non-public data for both training and evaluation, meaning that their results cannot be easily compared against each other.

As part of this work, we are releasing contextual biasing lists to accompany the Earnings21 dataset~\cite{del2021earnings}, creating a public benchmark for this task. We present baseline results on this benchmark using an open-source end-to-end ASR model from the WeNet toolkit~\cite{zhang2021wenet} trained on the GigaSpeech dataset~\cite{chen2021gigaspeech}. 

Previous work on contextual ASR is split between ``deep biasing" (modifications to ASR training so that context lists are a secondary input to the ASR model itself) and ``shallow fusion" (context lists used during decoding to bias the output). We will focus here on shallow fusion approaches that can be easily added onto an existing ASR model without retraining. We show results for shallow fusion applied to two decoding algorithms built into the WeNet toolkit, CTC prefix beam search and WFST decoding. 

Our baseline results confirm observations from prior work~\cite{zhao2019shallow,le2020g2g,le2021deep} that end-to-end models struggle in particular with words that are rarely or never seen during training, and that shallow fusion biasing alone does not adequately address this problem. We propose an alternate spelling prediction model that improves recall of rare words by 34.7\% relative and of out-of-vocabulary (OOV) words by 97.2\% relative, compared to contextual biasing without alternate spellings. Our model is conceptually similar to ones used in \cite{le2020g2g, huang2020class, le2021contextualized, le2021deep}, but is simpler to implement as it does not rely on either a pronunciation dictionary or an existing text-to-speech (TTS) system. 


\section{Related Work}

While prior work on contextual ASR includes both deep biasing (see \cite{pundak2018deep, bruguier2019phoebe, chen2019joint, alon2019contextual, jain2020contextual, sun2021tree}) and shallow fusion approaches, we chose to focus here on shallow fusion methods that can be applied to any existing ASR model without retraining. 

Perhaps the most common technique for contextual biasing in ASR is called on-the-fly (OTF) rescoring. This method was originally applied to hybrid ASR models~\cite{hall2015composition}. In that context, it involves composition of a weighted finite state transducer (WFST) representing the ASR model with a novel WFST representation of the bias terms, so that the weights on the bias terms are modified ``on the fly" at inference time. For end-to-end models, the bias terms are still compiled into a WFST representation, but the composition of that WFST with the ASR model happens at every decoding time-step - the likelihood of the current hypothesis according to the ASR model is combined with a score from the bias WFST. Several papers (\cite{williams2018contextual, chen2019end, zhao2019shallow}) explore versions of this technique. In particular, \cite{zhao2019shallow} demonstrates the importance of weight pushing to incorporate bias scores at each decoding time-step rather than only at word boundaries and of adding bias scores to all hypotheses before pruning the beam. 

Poor recognition of rare and out-of-vocabulary words even with contextual biasing has been repeatedly noted in the literature~\cite{zhao2019shallow, le2020g2g, le2021deep}. Several previous papers have introduced methods to mitigate this problem. These methods all use either TTS systems or pronunciation dictionaries. \cite{zhao2019shallow} uses both TTS and pronunciation dictionaries to augment their ASR training data and improve rare-word recognition in the core ASR model. TTS is used to synthesize audio for a variety of proper nouns; a pronunciation dictionary is used for ``transcript fuzzing" where proper nouns in the ASR training data are randomly replaced with other words in the lexicon that have the same pronunciation. 

A line of research by Le et al.~\cite{le2020g2g,le2021deep,le2021contextualized} uses TTS to generate alternate spellings for bias terms. Their method, which they call grapheme-to-grapheme (G2G) in reference to the grapheme-to-phoneme (G2P) systems used with phonemic models, is described in \cite{le2020g2g}, where they test it with a graphemic hybrid ASR model. In the hybrid framework, alternate spellings become alternate ``pronunciations" that can be added to the model without retraining. In \cite{le2021deep}, the authors apply this technique to an end-to-end RNN-T model using OTF rescoring, where these alternate spellings are added to the bias WFST. 
\cite{huang2020class} explores the same method of adding alternate paths for bias terms, but generates these alternates with a pronunciation dictionary. 

In this paper, our approach is based on the same concept of generating alternate spellings for bias terms, but our alternates are generated using no additional resources beyond the ASR model and training data. Because we are using an open-source model, this guarantees that our results can be replicated by other researchers. 

\section{Methods}

\subsection{ASR Model}
The ASR model for this work is an end-to-end model trained with both CTC~\cite{graves2006connectionist} and attention-based~\cite{chan2015listen} losses using the WeNet toolkit~\cite{zhang2021wenet}. The model has two components - an encoder and an attention-based decoder. The CTC loss is computed on the outputs of a softmax layer placed after the encoder outputs; the attention-based loss uses both the encoder and decoder. The model uses subword outputs. 


At inference time, we follow a two-pass decoding procedure. The first pass uses the encoder to produce an n-best list; the second pass uses the attention decoder to rescore the n-best list and choose the final output. We consider two different versions of the first pass. In the first version, we use CTC prefix beam search~\cite{graves2006connectionist}. For the second version, we use a WFST decoder from the Kaldi~\cite{povey2011kaldi} hybrid ASR toolkit, treating the CTC portion of our model as a graphemic acoustic model and combining it with an n-gram language model (LM)~\footnote{More information on this can be found in the WeNet documentation: \burl{https://wenet.org.cn/wenet/lm.html}}. For simplicity, we use a unigram language model. WFST decoding also requires a lexicon, which in a phonemic hybrid model is the list of words that can be recognized, along with one or more pronunciations for each. Here, the words in the lexicon are all of the words in the ASR training data, and the ``pronunciations" are their subword tokenizations. 

\subsection{Contextual Biasing}

The two versions of first-pass decoding described in the previous section allow us to experiment with two different methods of implementing contextual biasing. The first is CTC prefix beam search with OTF rescoring, in which we perform beam search but add additional weight to hypotheses that contain contextual biasing terms. This is accomplished by compiling the biasing terms into a WFST representation where paths represent a bias term and the arcs along the path represent the subword tokens that make up the term. For every hypothesis, at each decoding time-step, we consider whether we can enter the WFST or continue along a path within the WFST, depending on the most recent decoded token. For an easily-replicable baseline, we use the implementation within the WeNet toolkit, which applies the bias weight at every subword unit\footnote{See \burl{https://wenet.org.cn/wenet/context.html}}. For this biased CTC beam search approach, we individually add every word found in the biasing list to the bias WFST. 

Our second contextual biasing approach is a novel method implemented on top of WFST decoding, in which we directly modify the LM using the biasing list. Specifically, we add any terms from the biasing list that were not in the original LM and set the LM weight of all biasing terms to a fixed value. We also add new terms and their associated subword sequences to the lexicon, then recompile the WFST. We call this approach simply ``biased WFST decoding." In this case, we add multi-word bias terms to the WFST as single units, as well as adding each individual word within these terms on its own. 

For simplicity, we use $\beta$ to refer to the biasing weight parameter for both approaches. However, these parameter values are not comparable across the decoding methods. For OTF rescoring with CTC beam search, $\beta$ represents additional probability mass added at each subword unit within a bias word. For biased WFST decoding, $\beta$ is a log-probability applied to an entire bias term. In both cases, we consider values of $\beta$ such that our bias weights are no longer valid probabilities ($\beta>1.0$ for biased CTC decoding, $\beta>0.0$ for biased WFST decoding) because we find that these values can work well in practice.


\subsection{Error-Based Alternate Spelling Prediction Model}

We introduce a novel alternate spelling prediction (ASP) model that can be used to improve the performance of shallow fusion contextual biasing approaches. The ASP model is a text-to-text, transformer-based, encoder-decoder model. The input to this model is a word or phrase, and the desired output is a prediction of how the ASR system might, mistakenly, recognize that term. For example, if the input is the name ``Blac Chyna," the ASP model should predict ``black china" because that is how the ASR system will recognize the phrase. 

Shallow fusion contextual biasing puts additional weight on hypotheses containing the subword sequences associated with bias terms. Using this ASP model, we can also add weight to subword sequences associated with the alternate terms the model suggests. These sequences can then be replaced by the original bias terms in the ASR output. With the bias term ``Blac Chyna," this would be equivalent to adding ``black china" to the bias list so that it is more likely to be recognized, then replacing it with ``Blac Chyna" before returning the ASR output. In practice, these alternate subword sequences can be associated with the original context term inside of the WFST representation so that this replacement happens automatically as part of decoding.

Training data for the ASP model can be generated from any ASR training data. First, we decode all of the ASR training audio using greedy CTC decoding. Next, we align these outputs to the references. From these alignments, we extract all of the error segments in the outputs; we then filter these to only include errors made on non-stopword words and to not include insertion or deletion errors. 


At inference time, we use beam search to produce an n-best list of alternate spellings for each bias term. We then filter these alternates to remove bad alternates that are likely to introduce false-positive errors. There are two types of alternates that need to be filtered out: poor matches and common words. We filter out poor matches based on the ASP model likelihood produced during decoding - if the decrease in likelihood from one hypothesized alternate to the next best is above a threshold, we stop adding alternates for the current term. For common words, we filter out any suggested alternate that appeared more than a certain number of times in the training data. The size of the n-best list, the likelihood difference threshold, and the word count that defines common words are all adjustable parameters. 

\section{Data, Models, and Evaluation}

\subsection{Data}

For this paper, we use the GigaSpeech corpus~\cite{chen2021gigaspeech} for all model training - the ASR model, the alternate spelling prediction model, and the language model for WFST decoding. The GigaSpeech corpus is the open-source corpus that best reflects our task of interest: general-purpose ASR that can be applied to a wide range of domains. 

All results are reported on the Earnings21~\cite{del2021earnings} test set. This dataset consists of earnings calls from a variety of public companies. They contain a range of proper nouns - including the names of companies, products, and executives - that need to be recognized correctly for the transcripts to be comprehensible to a reader or useful for downstream processing.

For contextual biasing, we first created a ``best case scenario" list comprised of words and phrases contained in the Earnings21 references. These words and phrases were manually curated from the list of PERSON and ORG entities automatically tagged by spaCy~\cite{spacy2}. In real-world scenarios, biasing lists will also contain terms that are not found in the audio. We created a ``distractor" list by taking the original list and adding distractor items - names of Fortune 500 companies and famous CEOs. The best case list contains 1013 entries and the distractor list contains 1782 entries. Examples include ``Nokia" and ``Suzanne Sitherwood" from the best case list and ``Pepsico" and ``Elon Musk" from the distractor list. Both lists can be found in the Earnings21 repository~\footnote{\burl{https://github.com/revdotcom/speech-datasets/tree/main/earnings21}}. 

Unlike many ASR test sets, Earnings21 is not segmented into utterances. For inference, we segment each audio file into three minute chunks, and run the ASR model over each one separately. We append the ASR outputs for each audio file, and score those outputs against the reference for the whole file, using the  fstalign tool~\footnote{\burl{https://github.com/revdotcom/fstalign}}. 

\subsection{Evaluation Metrics}
We report word error rate (WER) for all experiments, but we find that contextual biasing has a minimal impact on overall WER for this use-case because the biasing terms are a small percentage of the overall words in the test set. 
In addition to WER, we report recall performance on biasing terms - we believe that this metric best captures the user's experience of the ASR system's ability to recognize these terms. We calculate this recall metric over 4 different categories of biasing terms: words, phrases, rare words, and OOV words. ``Words" refers to every non-stopword word that appears in the biasing list, even if it appears in the list as part of a multi-word phrase. ``Phrases" refers to multi-word phrases in the biasing list, when they appear in their complete form in the reference. Rare words are words seen less than 100 times in the ASR training data; OOV words are words never seen in the ASR training data.


\subsection{ASR Model}
We use an open-source WeNet ASR model~\footnote{Conformer bidecoder: \burl{https://github.com/wenet-e2e/wenet/tree/main/examples/gigaspeech/s0/}} for all experiments. It is a conformer-based model with 12 encoder layers and a bidirectional decoder with 3 layers in each direction. At inference time, we only use the right-to-left decoder. We use a chunk size of 512, a CTC weight of 0.1, and a rescoring weight of 1.0. The model uses an output vocabulary of 5000 subword units, learned from the training text using a unigram model~\cite{kudo2018subword}. 

For WFST decoding, we use a 1-gram language model trained on the text side of the Gigaspeech ASR training data. The base lexicon includes every word in the ASR training data. We use an acoustic scale of $8$ for all WFST experiments. 

For contextual biasing experiments, we report results as a function of the bias weight, $\beta$. For OTF rescoring with CTC, the bias weight is the amount of weight added at each subword unit during decoding of a bias term, and non-bias words recieve a weight of $0$. For biased WFST decoding, the bias weight is the LM score that we assign to bias terms; non-bias words have variable scores according to the LM, which are all strictly $<0$. 

\subsection{Alternate Spelling Prediction Model}
The alternate spelling prediction model is trained on a total of 673,366 reference/error pairs. The majority of this training data (73\%) has one word for both source and target. Only 3\% has more than two words on the source side. 

We used the OpenNMT toolkit~\footnote{\burl{https://github.com/OpenNMT/OpenNMT-py}} for our ASP model. The model is a transformer encoder-decoder model with two layers each in the encoder and decoder, two attention heads per layer, and 512 units per layer, for a total of 13.6M parameters. We found that that this ASP model was most effective with character tokenization on the input and subword tokenization on the output. The output subword tokenization is the same as the tokenization used for the ASR model. 

For all ASP experiments, we use an n-best list of five alternates. These are filtered based on a log-likelihood difference threshold of 1.0 and with common words defined as words that appear more than 1000 times in the ASR training data.

\section{Results and Discussion}

\subsection{Earnings21 Benchmark Baselines}

\begin{table}[t]
    \centering
    \begin{tabular}{|l|c|c|c|c|c|c|}
            \hline
        \multirow{2}{*}{} & 
        \multirow{2}{*}{$\beta$} & 
        \multirow{2}{*}{WER} &  
        \multicolumn{4}{c|}{Bias Recall} \\
        \cline{4-7}
        & & & Word & Phrase & Rare & OOV  \\
        \hline
        CTC & 0 & 13.23 & 85.4 & 60.7 & 22.5 & 5.9 \\
        & 0.5 & \textbf{13.19} & 86.6 & 64.1 & 28.3 & 10.5 \\
        & 1.0 & 13.27 & 87.5 & 66.2 & 33.9 & 13.9 \\
        & 1.5 & 13.38 & 88.3 & 69.1 & 38.4 & 17.3 \\
        & 2.0 & 13.64 & \textbf{88.9} & \textbf{70.8} & \textbf{41.8} & \textbf{20.8} \\
        \hline
        WFST & N/A & 13.10 & 86.2 & 62.9 & 25.7 & 0.0 \\
        & 2 & \textbf{12.99} & 87.9 & 70.9 & 41.2 & 21.9 \\
        & 4 & \textbf{12.99} & 88.3 & 72.4 & 43.9 & 23.0 \\
        & 8 & 13.05 & 88.7 & 74.2 & 48.0 & 26.4 \\
        & 12 & 13.21 & \textbf{89.0} & \textbf{75.8} & \textbf{51.4} & \textbf{28.9} \\
        \hline
    \end{tabular}
    \caption{Earnings21 baseline results. Note that the bias weight, $\beta$, is not comparable across the two contextual biasing approaches.}
    \label{tab:baseline}
\end{table}
Table \ref{tab:baseline} shows baseline results that we hope will serve as a foundation for future contextual biasing research. Comparing the two decoding methods without biasing, we can see that first-pass WFST decoding with a unigram LM produces slightly better WER than first-pass CTC beam search. The recall of phrases and of words seen during ASR training is higher with WFST decoding, but the language model prevents recognition of any OOV words. However, CTC decoding is also not very effective at recognizing OOV words - it only finds 5.9\% of the OOV words from the bias list, and only 7.6\% of the novel words it generates are correct.


In all cases, contextual biasing improves the recall of all categories of bias terms. 
With both versions of biased decoding, using a larger bias weight increases recall while also increasing WER. This is a consistent pattern seen throughout the results in this paper: increased recall of bias terms comes with additional false-positive errors that can cause significant WER increases at high enough weights. The acceptable trade-off between overall WER increases and improved recall of bias terms is application-dependent and easily adjusted based on the weight parameter for either biased decoding method.

Using the distractor list consistently increases WER, but by very small amounts. The difference in WER between runs with and without distractors gets larger with higher bias weights. With biased WFST decoding with $\beta=2$, the WER increases from 12.99 to 13.02 when the distractor list is included. With $\beta=8$, it goes from 13.05 to 13.12. 


\subsection{Alternate Spelling Prediction Models}

\begin{table}[]
    \centering
    \begin{tabular}{c|c|c|c|c}
        Term & & Alt. 1 & Alt. 2 & Alt. 3 \\
        \hline
        GAYLE & & GAIL & JAIL & GALE \\
        WAZE & & WAYS & WAS & WAZ \\
        NIRJANA & & NIRJANNA & NIJANA & NERJANA \\
        \hline
        AHYIANA & & AIANANA & AIANNA & AIANA \\
        HUSKAR & & HUSCAR & HUSAR & HUSKER 
    \end{tabular}
    \caption{Alternates generated by the ASP model. The first three words were in the model's training data, the last two words were unseen.} 
    \label{tab:g2g_alternates}
\end{table}

Table \ref{tab:g2g_alternates} shows examples of the types of alternates our ASP model produces. The first three rows are words that were in the training data for this model.
For ``GAYLE" and ``WAZE," the most likely alternate produced by the model matches what was in the training data. The suggestion of ``WAS" as an alternate for ``WAZE," while reasonable, is a clear example of the types of common words that need to be filtered out - the WER would be very high if we replaced every instance of ``WAS" in our ASR output with ``WAZE". During training, the model was shown ``NURJANA" as an alternate for ``NIRJANA" - we can see in this table that the model has not memorized the training data. The ASP model did not see the words in the last two rows of Table \ref{tab:g2g_alternates} during training, but it is still able to generate reasonable alternatives for them.

\begin{table}[b]
    \centering
    \begin{tabular}{|c|c|c|c|c|c|}
        \hline
        \multirow{2}{*}{$\beta$} & 
        \multirow{2}{*}{WER} &  
        \multicolumn{4}{c|}{Bias Recall} \\
        \cline{3-6}
        & & Word & Phrase & Rare & OOV  \\
        \hline
        -4.0 & \textbf{12.97} & 88.4 & 73.3 & 48.5 & 37.1 \\
        -2.0 & 13.07 & 88.7 & 74.4 & 50.7 & 39.4 \\
        0.0 & 13.08 & 88.9 & 75.5 & 53.3 & 41.9 \\
        2.0 & 13.14 & 89.2 & 76.3 & 55.5 & 43.2 \\
        4.0 & 13.20 & 89.3 & 77.2 & 57.3 & 44.5 \\
        8.0 & 13.43 & \textbf{89.5} & \textbf{78.5} & \textbf{60.0} & \textbf{47.1} \\
        \hline
    \end{tabular}
    \caption{Results of biased WFST decoding with alternate spellings, as a function of bias weight, $beta$.}
    \label{tab:g2g_wfst_bias_weight}
\end{table}

Because WFST decoding showed better results in the previous section, we only show alternate spellings results for biased WFST decoding here. 
Table \ref{tab:g2g_wfst_bias_weight} has these results, as a function of the bias weight. As in Table \ref{tab:baseline}, Table \ref{tab:g2g_wfst_bias_weight} shows a clear trade-off between bias term recall and WER, this time with a more pronounced increase in WER as the bias weight increases. With a bias weight of $-2.0$ and the ASP model, we can achieve $39.4\%$ recall of OOV words and $50.7\%$ recall of rare words while decreasing WER slightly from the WFST decoding baseline. Conversely, if we are willing to accept a small increase in WER (up to $13.20\%$, comparable to the CTC prefix beam search baseline), we can get $44.5\%$ recall of OOV words and $57.3\%$ recall of rare words with a bias weight of $4.0$. 

By comparing the results in Table\ref{tab:g2g_wfst_bias_weight} to those in Table \ref{tab:baseline}, we can see that adding alternate spellings has significantly improved the recall of rare and OOV words while also increasing WER. For example, with $\beta=2$, the ASP model improves phrase recall by 7.6\% relative, rare word recall by 34.7\% relative, and OOV word recall by 97.2\% relative over biased WFST decoding without alternate spellings. However, the WER has also increased, from 12.99\% to 13.14\%.

Because of the WER increases when alternate spellings are used, we can also understand the impact of the ASP model by comparing models that achieve approximately the same WER rather than those that use the same bias weight. For example, we can compare a bias weight of 12 with no alternates to a bias weight of 4 with alternates. In that case, alternate spellings improve phrase recall by 1.8\% relative, rare word recall by 11.5\% relative, and OOV word recall by 54.0\% relative.



There are a few reasons why many bias terms do not get recognized even with alternate spellings. The most common issue is alternate spellings that do not accurately capture the sound of the word. For example, the bias term ``ACELRX" is recognized by the ASR model as ``EXCEL X" but the top alternates are ``AERX" and ``ACELLS." We expect that many of these errors could be fixed by adding more ASP training data or combining our technique with a pronunciation-based method like \cite{huang2020class}. Another common category of error is bias terms that sound like common words, like ``GALLERI." The ASP model correctly suggests ``GALLERY," but that alternate is filtered out because it meets our threshold for a common word that cannot be replaced. Ideally, there would be a method for determining whether a common word could be replaced by a bias term based on context, instead of our current all-or-nothing approach. This method could potentially be part of the rescoring pass.

Lastly, we look at the impact of distractor items when using ASP. With WFST decoding, a bias weight of 12.0 without alternates got a similar WER to a bias weight of 4.0 with alternates (13.21\% vs. 13.20\%). When we use the distractor list, these WERs increase to 13.35\% and 13.31\%, respectively. The ASP model slightly lessens the impact of distractors, perhaps because it allows us to use a lower bias weight. 

\section{Conclusions}
This paper makes two contributions. First, we release standardized contextual biasing lists for a public dataset along with baseline results that can be easily replicated with an open-source toolkit and available pretrained model. Second, we present a novel alternate spelling prediction model and demonstrate its usefulness in improving contextual recognition of rare and unseen words. Combined with biased WFST decoding, this model achieves a 34.7\% relative increase in rare-word recall and a 97.2\% relative increase in OOV recall.

While the results using our ASP model are promising, we hope that the release of an open benchmark will inspire additional research on this task. In future work, we plan to explore the use of alternate spellings from multiple sources and to combine these alternate spellings with deep biasing approaches to contextual ASR. Additionally, we plan to look into methods for incorporating contextual biasing into second-pass rescoring.

\bibliographystyle{IEEEtran}

\bibliography{mybib}

\end{document}